\documentclass[11pt,a4paper]{article}
\usepackage[hyperref]{emnlp2020}
\usepackage{times}
\usepackage{latexsym}

\usepackage{graphicx}
\usepackage{amsmath}
\usepackage{amsfonts}
\usepackage{amssymb}
\usepackage{color}
\usepackage{multirow}
\usepackage{adjustbox}
\usepackage[english]{babel} 

\usepackage{microtype}
\usepackage{arydshln}

\aclfinalcopy 

\title{AGIF: An Adaptive Graph-Interactive Framework for Joint Multiple Intent Detection and Slot Filling}

\author{Libo Qin, Xiao Xu, Wanxiang Che\thanks{\ \  Email corresponding.}, Ting Liu \\
	Research Center for Social Computing and Information Retrieval \\
	Harbin Institute of Technology, China \\
	{\tt \{lbqin,xxu,car,tliu\}@ir.hit.edu.cn}	
}

\date{}
\begin{document}
\maketitle
\begin{abstract}
In real-world scenarios, users usually have multiple intents in the same utterance.
Unfortunately, most spoken language understanding (SLU) models either mainly focused on the single intent scenario, or simply incorporated an overall intent context vector for all tokens, ignoring the fine-grained multiple intents information integration for token-level slot prediction.
In this paper, we propose an \textbf{A}daptive \textbf{G}raph-\textbf{I}nteractive \textbf{F}ramework (AGIF) for joint multiple intent detection and slot filling, where we introduce an intent-slot graph interaction layer to model the strong correlation between the slot and intents.
Such an interaction layer is applied to each token adaptively, which has the advantage to automatically extract the relevant intents information, making a fine-grained intent information integration for the token-level slot prediction.
 Experimental results on three multi-intent datasets show that our framework obtains substantial improvement and achieves the state-of-the-art performance.
In addition, our framework achieves new state-of-the-art performance on two single-intent datasets.
\end{abstract}

\section{Introduction}
\label{Introduction}
\begin{figure*}[t]
	\centering
	\includegraphics[width=1.0\textwidth]{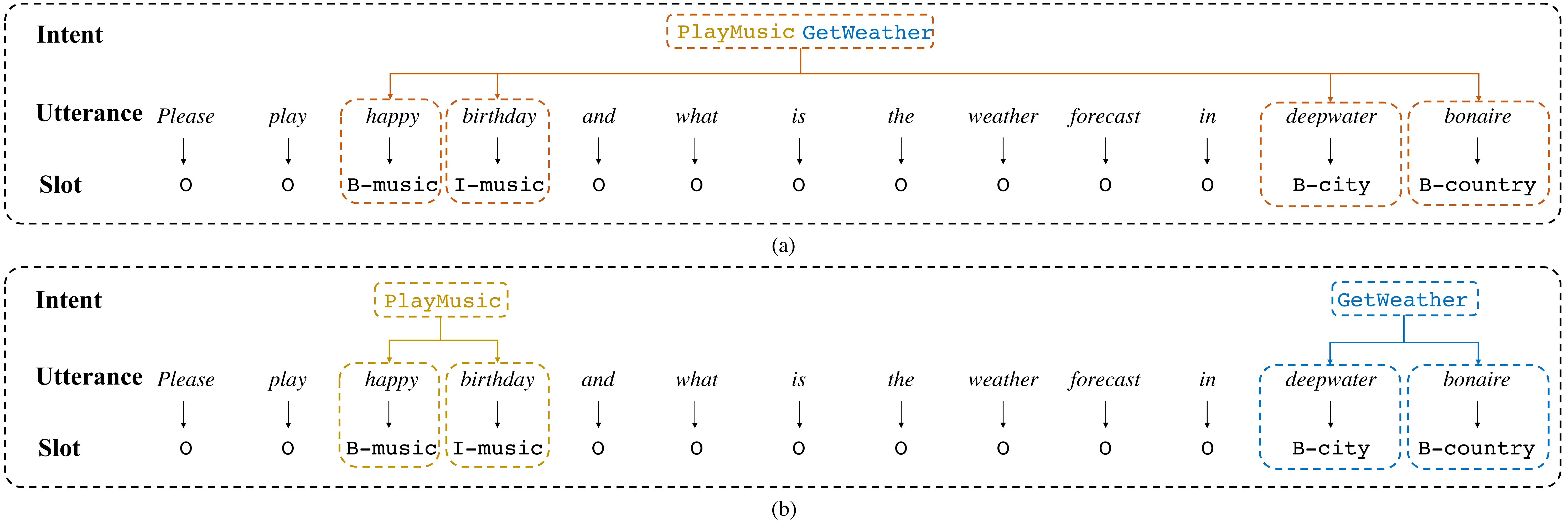}
	\caption{Prior model simply treat multiple intents as an overall intent information (a) vs. our fine-grained multiple intents integration method (b).}
	\label{fig:example}
\end{figure*}
Spoken language understanding (SLU)~\citep{young2013pomdp} is a core component of task-oriented dialog systems.
It consists of two typical subtasks, intent detection and slot filling~\citep{tur2011spoken}.
Take the utterance ``\textit{Please play {happy birthday}}'' for example, 
the intent detection can be seen as a classification task to classify the intent label (\emph{i.e.}, \texttt{PlayMusic}) while the slot filling can be treated as a sequence labeling task to predict the slot label sequence (\emph{i.e.}, \texttt{O}, \texttt{O}, \texttt{B-music}, \texttt{I-music}).
Dominant SLU systems in the literature~\citep{goo2018slot,li2018self,e-etal-2019-novel,liu2019cm,qin-etal-2019-stack}
adopt joint models to model the relation between the two tasks, which is a direction we follow.

Though achieving promising performances, most prior work only focus on the simple single intent scenario.
Their models are trained based on the assumption that each utterance only has one single intent.
Actually, users usually express multiple intents in an utterance and \citet{gangadharaiah2019joint} shows that 52\% of examples are multi-intent in the amazon internal dataset.
Nevertheless, the existing trained single intent SLU models fail to effectively handle the multi-intent settings with the original network structure.
Ideally, when an SLU system meets an utterance with multiple intents, as shown in Figure~\ref{fig:example}(a), the model should directly detect its all intents (\texttt{PlayMusic} and \texttt{GetWeather}).
Hence, it is important to consider multi-intent SLU.

Unlike the prior single intent SLU model which can simply leverage the utterance's single intent to guide slot prediction \cite{goo2018slot,qin-etal-2019-stack}, multi-intent SLU faces to multiple intents and presents a unique challenge that is worth studying: how to effectively incorporate multiple intents information to lead the slot prediction.
To this end, \citet{gangadharaiah2019joint} first explored the multi-task framework with the slot-gated mechanism \citep{goo2018slot} for joint multiple intent detection and slot filling.
Their model incorporated intent information by simply treating an intent context vector as multiple intents information.
While this is a direct method for incorporating multiple intents information, it does not offer fine-grained intent information integration for token-level slot filling in the sense that each token is guided with the same complex intents information, which is shown in Figure~\ref{fig:example}(a).
In addition, providing the same intent information for all tokens may introduce ambiguity, where it's hard for each token to capture the related intent information.
As shown in Figure~\ref{fig:example}(b), these tokens ``\textit{happy birthday}'' should focus on the intent ``\texttt{PlayMusic}'' while tokens ``\textit{deepwater bonaire}'' depend on the intent ``\texttt{GetWeather}''.
Thus, each token should focus on the corresponding intent and it's critical to make a fine-grained intent information integration for the token-level slot prediction.

In this paper, we propose an \textbf{A}daptive \textbf{G}raph-\textbf{I}nteractive \textbf{F}ramework (AGIF)  to address the aforementioned concern.
The core module is the proposed adaptive intent-slot graph interaction layer, which is constructed of each token's hidden state of slot filling decoder and embeddings of predicted multiple intents.
In this graph, each token's slot node directly connects all predicted intent nodes to explicitly build the correlation between slot and intents.
Such an interaction graph is applied to each token adaptively, which make each token has the ability to capture different relevant intent information
so that fine-grained multiple intents integration can be achieved.
In contrast to prior work simply incorporate multiple intents information statically where the same intents information is used for guiding all tokens, our intent-slot interaction graph is constructed adaptively with graph attention network over each token.
This encourages our model to automatically filter the irrelevant information and capture important intent at the token-level.

We first conduct experiments on the multi-intent benchmark dataset DSTC4~\citep{schuster-etal-2019-cross-lingual}.
Then, to verify the generalization of our framework, we empirically construct two large-scale multi-intent datasets MixATIS~\citep{hemphill1990atis} and MixSNIPS~\citep{coucke2018snips}.
The results of these experiments show the effectiveness of our framework by outperforming the current state-of-the-art method.
To the best of our knowledge, there are no public large-scale multiple intents datasets and we hope the release of it would push forward the research of multi-intent SLU. 
In addition, our framework achieves state-of-the-art performance on two public single-intent datasets including ATIS \citep{tur2011spoken} and SNIPS~\citep{coucke2018snips}, which further verifies the generalization of the proposed model.

To facilitate future research in this area, all datasets and codes are publicly available at \url{https://github.com/LooperXX/AGIF}.

\section{Approach}
\label{Approach}
\begin{figure*}[t]
	\centering
	\includegraphics[width=0.95\textwidth]{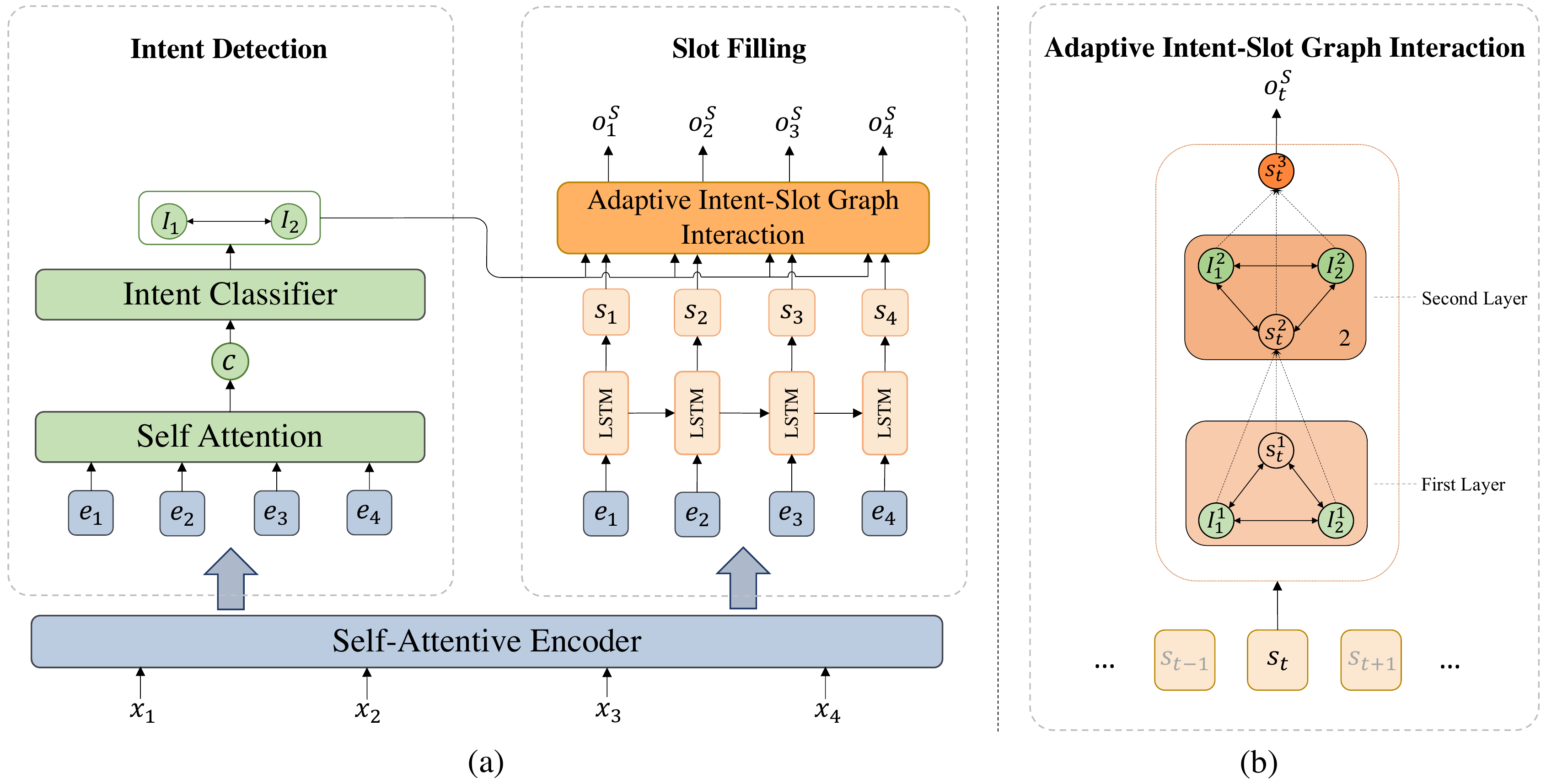}
	\caption{The overflow of model architecture (a) and adaptive intent-slot graph interaction module (b).}
	\label{fig:framework}
\end{figure*}
The architecture of our framework is demonstrated in Figure~\ref{fig:framework}, which consists of a shared encoder, an adaptive intent-slot graph interaction layer and two separate decoders.
First, the encoder ($\S\ref{sec:encoder}$) uses a shared self-attentive encoder to represent an utterance, which can grasp the shared information between intent detection and slot filling.
Then, the intent detection decoder ($\S\ref{sec:intent_decoder}$) performs the multi-label classification to detect multiple intents.
Finally, we introduce the adaptive intent-slot graph interaction layer ($\S\ref{sec:slot_filling_decoder}$) to explicitly leverage the multiple intents information for guiding slot prediction.
Both intent detection and slot filling are optimized simultaneously via a multi-task learning scheme.

\subsection{Self-Attentive Encoder}\label{sec:encoder}
In the self-attentive encoder, following \citet{qin-etal-2019-stack}, we use BiLSTM with the self-attention mechanism to 
leverage both advantages of temporal features within word orders and contextual information.

\paragraph{Bidirectional LSTM} 
A bidirectional LSTM (BiLSTM)~\citep{hochreiter1997long} consists of two LSTM layers.
For the input sequence $\{{\boldsymbol{x}}_{1}, {\boldsymbol{x}}_{2}, \ldots, {\boldsymbol{x}}_{T}\}$ ($T$ is the number of tokens in the input utterance), the BiLSTM reads it 
forwardly from $\boldsymbol{x}_{1}$ to $\boldsymbol{x}_{T}$ and backwardly from $\boldsymbol{x}_{T}$ to $\boldsymbol{x}_{1}$ to produce a series of context-sensitive hidden states $\boldsymbol{H} = \{\boldsymbol{h}_{1}, \boldsymbol{h}_2, \ldots, \boldsymbol{h}_{T}\}$.

\paragraph{Self-Attention}  We follow \citet{NIPS2017_7181} to use a self-attention mechanism over word embedding to capture context-aware features. 
We first map the matrix of input vectors $\boldsymbol{X} \in \mathbb R^{T \times d}$ ($d$ represents the mapped dimension) to queries $\boldsymbol{Q}$, keys $\boldsymbol{K}$ and values $\boldsymbol{V}$ matrices by using different linear projections parameters $\boldsymbol{W}_{q}$,  $\boldsymbol{W}_{k}$, $\boldsymbol{W}_{v}$.
Attention weight is computed by dot product between $\boldsymbol{Q}$, $\boldsymbol{K}$ and the self-attention output $\boldsymbol{A} \in \mathbb R^{T \times d}$ is a weighted sum of values:
\begin{equation}
\boldsymbol{A} 
= \operatorname{softmax}\left( \frac{\boldsymbol{Q}\boldsymbol{K}^{\top}}{\sqrt{d_k}}\right){\boldsymbol{V}} \,,
\end{equation}
where $d_{k}$ denotes the dimension of keys.

We concatenate these two representations as 
the final encoding representation:
\begin{equation}
{\boldsymbol{E}} =[{\boldsymbol{H}} \mathop{||} {\boldsymbol{A}}] \,,
\end{equation}
where $\boldsymbol{E}$ = $\{\boldsymbol{e}_{1},\ldots,\boldsymbol{e}_{T}\} \in \mathbb{R}^{T\times 2d}$ and $\mathop{||}$ is concatenation operation.

\subsection{Intent Detection Decoder} \label{sec:intent_decoder}
We follow \citet{gangadharaiah2019joint} to perform multiple intent detection as the multi-label classification problem. 
We compute the utterance context vector over $\boldsymbol{E}$ = $\{\boldsymbol{e}_{1},\ldots,\boldsymbol{e}_{T}\} \in \mathbb{R}^{T\times 2d}$.
In our case, we use a self-attention module~\citep{P18-1135,goo2018slot} to capture relevant context:
\begin{align}
\label{eq:selfattnscore}
p_t &= \operatorname{softmax} (\boldsymbol{w}_e~\boldsymbol{e}_t + \boldsymbol{b}) \,, \\
\boldsymbol{c} &= \sum_t p_t \boldsymbol{e}_t \,, 
\end{align}
where $\boldsymbol{w}_e$ $\in \mathbb R^{1 \times2d}$ is the trainable parameters, $p_{t}$ is corresponding normalized self-attention score.

$\boldsymbol{c}$ is the weighted sum of each element $\boldsymbol{e}_{t}$ and utilized for intent detection:
\begin{eqnarray}
\label{eq:intent-detection}
\boldsymbol{y}^{I} \!=\! \sigma(\boldsymbol{W}_{i}(\operatorname{LeakyReLU}(\boldsymbol{W}_{c}~\boldsymbol{c} \!+\! \boldsymbol{b}_c))\!+\! \boldsymbol{b}_i) \,,
\end{eqnarray}
where $\boldsymbol{W}_{i}, \boldsymbol{W}_{c}$ are trainable parameters of the intent decoder, ${\boldsymbol{y}}^{I} = \{y_{1}^{I}, \ldots, y_{N_{I}}^{I}\}$ is the intent output of the utterance and $N_I$ is the number of single intent labels. $\sigma$ represents the sigmoid activation function.

During inference, we predict intents $\boldsymbol{I}$ = \{$I_{1},\dots,I_{n}\}$ and $I_{i}$ represents probability $y_{I_{i}}^{I}$ greater than $t_{u}$, where 0 $<$ $t_{u}$ $<$ 1.0 is a hyperparameter tuned using the validation set.\footnote{In our experiments, we set $t_{u}$ as 0.5.} 
For example, if the ${\boldsymbol{y}}^{I} = \{0.9, 0.3, 0.6, 0.7, 0.2\}$ and the $t_{u}$ is 0.5, we predict intents $\boldsymbol{I}$ = $\{1, 3, 4\}$.

\subsection{Adaptive Intent-Slot Graph Interaction for Slot Filling} \label{sec:slot_filling_decoder}
In this paper, one of the core contribution is adaptively leveraging multiple intents to guide the slot prediction, encouraging each token to capture the corresponding relevant intent information.
In particular, we adopt the graph attention network (GAT) \citep{velivckovic2017graph} to model the interaction between intents and slot at the token-level.

In this section, we first describe the vanilla graph attention network. Then, we show how to directly leverage multiple intents information for slot prediction with the adaptive intent-slot graph interaction layer.

\paragraph{Vanilla Graph Attention Network}
For a given graph with $N$ nodes, one-layer GAT take the initial node features $\tilde{\boldsymbol{H}} = \{\tilde{\boldsymbol{h}}_{1},\ldots, \tilde{\boldsymbol{h}}_{N}\}, \tilde{\boldsymbol{h}}_{n} \in \mathbb{R}^{F}$ as input, aiming at producing more abstract representation, $\tilde{\boldsymbol{H}}^\prime = \{\tilde{\boldsymbol{h}}^{\prime}_{1},\ldots,\tilde{\boldsymbol{h}}^{\prime}_{N}\}, \tilde{\boldsymbol{h}}^{\prime}_{n} \in \mathbb{R}^{F^\prime}$, as its output. The graph attention operated on the node representation can be written as:
\begin{equation}\label{eq:gat}
\begin{aligned} 
\mathcal{F}(\tilde{\boldsymbol{h}}_{i}, \tilde{\boldsymbol{h}}_{j}) &= \operatorname{LeakyReLU}\left(\mathbf{a}^\top[\boldsymbol{W}_h\tilde{\boldsymbol{h}}_i\|\boldsymbol{W}_h\tilde{\boldsymbol{h}}_j]\right) \,,\\
\alpha_{ij} &= \frac{\exp(\mathcal{F}(\tilde{\boldsymbol{h}}_{i}, \tilde{\boldsymbol{h}}_{j}))}{\sum_{j^\prime \in \mathcal{N}_i} \exp{(\mathcal{F}(\tilde{\boldsymbol{h}}_{i}, \tilde{\boldsymbol{h}}_{j^\prime}))}} \,, \\
\tilde{\boldsymbol{h}}^{\prime}_i &=  \sigma\big(\sum_{j \in \mathcal{N}_i} \alpha_{ij} \boldsymbol{W}_h\tilde{\boldsymbol{h}}_{j}\big) \,,
\end{aligned}
\end{equation}
where $\mathcal{N}_i$ is the first-order neighbors of node $i$ (including $i$) in the graph,  $\boldsymbol{W}_h \in \mathbb{R}^{F^\prime \times F}$ and $\mathbf{a} \in \mathbb{R}^{2F^\prime}$ is the trainable weight matrix, $\alpha_{ij}$ is the normalized attention weight denoting the importance of each $\tilde{\boldsymbol{h}}_{j}$ to $\tilde{\boldsymbol{h}}_{i}$ and $\sigma$ represents the nonlinearity activation function. 

GAT inject the graph structure into the mechanism by performing \textit{masked attention}, i.e, GAT only compute $\mathcal{F}(\tilde{\boldsymbol{h}}_{i}, \tilde{\boldsymbol{h}}_{j})$ for nodes $j \in \mathcal{N}_i$.
To stabilize the learning process of self-attention, GAT extend the above mechanism to employ \textit{multi-head attention} from~\citet{NIPS2017_7181}: 
\begin{equation} \label{eq:multi-head}
	\tilde{\boldsymbol{h}}^{\prime}_i =  \mathop{||}_{k=1}^{K} \sigma\big(\sum_{j \in \mathcal{N}_i} \alpha_{ij}^k \boldsymbol{W}_h^k \tilde{\boldsymbol{h}}_{j}\big) \,,
\end{equation}
where $\alpha_{ij}^k$ is the normalized attention weight computed by the $k$-th function $\mathcal{F}_k$, $\mathop{||}$ is concatenation operation and $K$ is the number of heads. Thus, the output $\tilde{\boldsymbol{h}}^{\prime}_{n}$ will consists of $KF^\prime$ features in the middle layers and the final prediction layer will employ averaging instead of concatenation to get the final prediction results.

\paragraph{Adaptive Intent-Slot Graph Interaction for Slot Prediction}
We use a unidirectional LSTM as the slot filling decoder.
At each decoding step $t$, the decoder state ${\boldsymbol{s}}_{t}$ is 
calculated by previous decoder state ${\boldsymbol{s}}_{t-1}$, the previous 
emitted slot label distribution ${\boldsymbol{y}}_{t-1}^{S}$ 
and the aligned encoder hidden state ${\boldsymbol{e}}_{t}$:
\begin{equation}
{\boldsymbol{s}}_{t} = \operatorname{LSTM} \left({\boldsymbol{s}}_{t-1}, {\boldsymbol{y}}_{t-1}^{S}, {\boldsymbol{e}}_{t} \right).
\end{equation}

Instead of directly utilizing the $\boldsymbol{s}_{t}$ to predict the slot label, we build a graphic structure named adaptive intent-slot graph interaction to explicitly leverage multiple intents information to guide the $t$-th slot prediction.
In this graph, the slot hidden state at $t$ time step is $\boldsymbol{s}_{t}$ and predicted multiple intents information ${\boldsymbol{I}} = \{I_{1}, \ldots, I_{n}\}$, where $n$ denotes the number of predicted intents, are used as the initialized representations at $t$ time step $\tilde{\boldsymbol{H}}^{[0, t]} =\{\boldsymbol{s}_{t}, \phi^{emb}(I_{1}), \ldots, \phi^{emb}(I_{n})\} \in \mathbb{R}^{(n+1) \times d}$, where $d$ represents the dimension of vertices representation and $\phi^{emb}(\cdot)$ represents the embedding matrix of intents. 
In addition, the predicted intents are connected to each other to consider their mutual interaction because all of them express the same utterance's intent.

For convenience, we use $\tilde{\boldsymbol{h}}_{i}^{[l, t]}$ to represent node $i$ in the $l$-th layer of the graph consisting of the decoder state node and predicted intent nodes at $t$ time step. $\tilde{\boldsymbol{h}}_{0}^{[l, t]}$ is the slot hidden state representation in the $l$-th layer.
To explicitly leverage the multiple intents information, the slot hidden state node is directly connected to all predicted intents and the slot node representation in the $l$-th layer can be calculated as:
\begin{equation}
	\label{eq:aggregation}
	\tilde{\boldsymbol{h}}_{i}^{[l, t]} = \sigma\big(\sum_{j \in \mathcal{N}_i} \alpha_{ij}^{[l, t]} \boldsymbol{W}_h^{[l]} \tilde{\boldsymbol{h}}_{j}^{[l-1, t]}\big), 
\end{equation}
where $\mathcal{N}_{i}$ represents the first-order neighbors of node $i$, \emph{i.e.}, the decoder state node and the predicted intent nodes, and the update process of all node representations can be calculated by Equation~\ref{eq:gat},~\ref{eq:multi-head} and~\ref{eq:aggregation}.

With $L$-layer adaptive intent-slot graph interaction, we obtain the final slot hidden state representation $\tilde{\boldsymbol{h}}_{0}^{[L, t]}$ at $t$ time step,
which adaptively capture important intents information at token-level. The representation $\tilde{\boldsymbol{h}}_{0}^{[L, t]}$ is utilized for slot filling:
\begin{align}
{\boldsymbol{y}}_{t}^{S} &= \operatorname{softmax} \left({\boldsymbol{W}}_{s}\tilde{\boldsymbol{h}}_{0}^{[L, t]}\right),\\
{o}_{t}^{S} &= \arg \max({\boldsymbol{y}}_{t}^{S}),
\end{align}
where ${o}_{t}^{S}$ is the predicted slot label of the $t$-th word in the utterance.

\subsection{Multi-Task Training}
Following \newcite{qin2020dcr}, we adopt a joint model to consider the two tasks and update parameters by joint optimizing.
The intent detection objective is:
\begin{equation}
	\mathcal{L}_{1} \triangleq -\sum_{k=1}^{N_I} \left(\hat{y}_{k}^{I} \log \left({y}_{k}^{I}\right)+\left(1-\hat{y}_{k}^{I}\right) \log \left(1-{y}_{k}^{I}\right)\right) \,.
\end{equation}

Similarly, the slot filling task objective is defined as: 
\begin{equation}
\mathcal{L}_{2} \triangleq - \sum_{i=1}^{M}\sum_{j=1}^{N_S}{{\hat{{y}}_{i}^{(j,S)}}\log \left({{y}}_{i}^{(j,S)}\right)},
\end{equation}
where $N_I$ is the number of single intent labels, $N_S$ is the number of slot labels and $M$ is the number of words in an utterance.

The final joint objective is formulated as:
\begin{equation}
\mathcal{L}=\alpha \mathcal{L}_{1} +(1-\alpha)  \mathcal{L}_{2},
\end{equation}
where $\alpha$ is hyper-parameter.

\section{Experiments}
\label{Experiments}

\begin{table*}[tbp]
	\centering
	\begin{adjustbox}{width=1.0\textwidth}
		\begin{tabular}{l|c|c|c|c|c|c|c|c}
			\hline
			\multirow{2}{*}{\textbf{Model}} & \multicolumn{4}{c|}{\textbf{MixATIS}} & \multicolumn{4}{c}{\textbf{MixSNIPS}} \\ \cline{2-9}
			& \textbf{Slot (F1)}      & \textbf{Intent (F1)}   & \textbf{Intent (Acc)}   & \textbf{Overall (Acc)}  & \textbf{Slot (F1)}      & \textbf{Intent (F1)}   & \textbf{Intent (Acc)}   & \textbf{Overall (Acc)} \\ \hline
			Attention BiRNN   & 86.6 &  -  & 71.6 & 38.7 & 89.4 & - & 94.1 & 62.2 \\
			Slot-Gated  & \textbf{88.1} &  -  & 65.7 & 38.9 & 87.8 & - & 96.0 & 56.5 \\
			Slot-gated Intent  & 86.7 &  -  & 66.2 & 39.6 & 87.9 & - & 94.2 & 57.6 \\
			Bi-Model         & 85.5 &  -  & 72.3 & 39.1 & 86.8 & - & 95.3 & 53.9 \\
			SF-ID             & 87.7 &  -  & 63.7 & 36.2 & 89.6 & - & 96.3 & 59.3 \\
			Stack-Propagation (\textit{concatenation})  & 86.6 &  -  & \textbf{76.0} & 42.8 & 93.9 & - & 96.4 & 75.5 \\
			Stack-Propagation (\textit{sigmoid-decoder}) & 87.4 &  79.0  & 71.9 & 41.0 & 93.2 & 97.6 & 94.6 & 71.9 \\
			\hdashline
			Joint Multiple ID-SF & 87.5 & 80.6 & 73.1 & 38.1 & 91.0 &  98.2  & 95.7 & 66.6  \\  \hline
			AGIF & \textbf{88.1} & \textbf{81.2*} & 75.8 & \textbf{44.5*} & \textbf{94.5*} & \textbf{98.6*} & \textbf{96.5*} & \textbf{76.4*}  \\ \hline
						
		\end{tabular}
	\end{adjustbox}
	\caption{Slot filling and intent detection results on two self-constructed multi-intent datasets. The numbers with * indicate that the improvement of our model over all the compared baselines is statistically significant with p \textless 0.05 under the t-test.}
	\label{tab:multi_intent_results}
\end{table*}

\subsection{Datasets}
\label{datasets}
\paragraph{Multiple Intent Datasets}
We conduct experiments on the benchmark DSTC4~\citep{kim2017fourth},
which is human-human multi-turn dialogues.
We adopt the same dataset partition in the DSTC4 main task and we regard its speech act attributes as intents.\footnote{The official DSTC4 pilot tasks' Handbook \url{http://www.colips.org/workshop/dstc4/DSTC4_pilot_tasks.pdf}}
It has 12,759 utterances for training, 4,812 utterances for validation and 7,848 utterances for testing.

To verify the generalization of the proposed model, we construct the multi-intent SLU dataset, {MixSNIPS}.
{MixSNIPS} dataset is collected from the Snips personal voice assistant \cite{coucke2018snips} by using conjunctions, \emph{e.g.}, ``and'', to connect sentences with different intents and ensure that the ratio of sentences has 1-3 intents is $[0.3,0.5,0.2]$. 
Finally, we get the 45,000 utterances for training, 2,500 utterances for validation and 2500 utterances for testing on the MixSNIPS dataset.
Similarly, we construct another multi-intent SLU dataset, MixATIS, from the {ATIS} dataset~\citep{hemphill1990atis}. There are 18,000 utterances for training, 1,000 utterances for validation and 1,000 utterances for testing. 
The constructed datasets have been released for future research.
\paragraph{Single Intent Datasets}
In addition, we also conduct experiments on two public benchmark single-intent datasets to validate the efficiency of our proposed model. One is the {ATIS} dataset~\citep{hemphill1990atis} and the other is  {SNIPS} dataset~\citep{coucke2018snips}, which 
are widely used as benchmark in SLU research. Both datasets follow the same format and partition as in \citet{goo2018slot} and \citet{qin-etal-2019-stack}. 

\subsection{Experimental Settings}
The self-attentive encoder hidden units is $256$ in all datasets. $\ell_2$ regularization is $1 \times 10^{-6}$ and dropout rate is $0.4$ for reducing overfitting. We use Adam~\citep{kingma2014adam} to optimize the parameters in our model and adopted the suggested hyper-parameters for optimization.
The graph layer number is $3$ for DSTC4 dataset and $2$ for the other datasets. For all the experiments, we select the model which works the best on the dev set and then evaluate it on the test set. 
All experiments are conducted at TITAN Xp and GeForce RTX 2080Ti. The epoch number is $50$ for MixSNIPS and $100$ for MixATIS and DSTC4.

\subsection{Baselines}
We first compare our model with the existing state-of-the-art multi-intent SLU baseline: \\ {\bf{Joint Multiple ID-SF}}.  \citet{gangadharaiah2019joint} proposes
a multi-task framework with the slot-gated mechanism for multiple intent detection and slot filling.

Then, we compare our framework with the existing state-of-the-art single-intent SLU:\\
1) {\bf{Attention BiRNN.}} \citet{liu2016attention} propose an alignment-based RNN with the attention mechanism, which implicitly learns the relationship between slot and intent. \\
2)   {\bf{Slot-Gated Atten.}} \citet{goo2018slot} proposes a slot-gated joint model to explicitly consider the correlation between slot filling and intent detection. \\
3)  {\bf{Bi-Model.}} \citet{wang2018bi} proposes the Bi-model to consider the cross-impact between the intent detection and slot filling. \\
4)  {\bf{SF-ID Network.}} \citet{haihong2019novel} proposes an SF-ID network to establish direct connections for the slot filling and intent detection to help them promote each other mutually. \\
5) {\bf{Stack-Propagation.}}  \citet{qin-etal-2019-stack} adopts a joint model with Stack-Propagation to capture the intent semantic knowledge and perform the token-level intent detection to further alleviate the error propagation.
This model achieves the state-of-the-art performance.

To enable single-intent SLU baselines can handle the multi-intent utterances,  we follow \citet{gangadharaiah2019joint} to connect them with $\#$ to get the single multi-intent label for a fair comparison, we name it as \textit{concatenation} version. To further verify the effectiveness of our framework, we change the state-of-the-art baseline \textit{Stack-Propagation} to directly predict the multi-intent label by changing the inten decoder with replacing softmax as sigmoid and using binary cross-entropy loss. We refer it as the \textit{sigmoid-decoder}.

For the \textit{Attention BiRNN}, \textit{Slot-Gated Atten}, \textit{SF-ID Network} and \textit{Stack-Propagation}, we run their official source code to obtain the results. For the \textit{Bi-Model} and \textit{Joint Multiple ID-SF}, we re-implemented the models and obtained the results on the same datasets because the original paper did not release their codes.

\subsection{Main Results}
\label{Results}

Following \citet{goo2018slot} and \citet{qin-etal-2019-stack}, we evaluate the performance of slot filling using F1 score, intent prediction using accuracy and macro F1 score, the sentence-level semantic frame parsing using overall accuracy which represents all metrics are right in an utterance. Table~\ref{tab:multi_intent_results} shows the experiment results of the proposed models on the MixATIS and MixSNIPS datasets. 

\begin{table}[h]
	\centering
	\begin{adjustbox}{width=0.48\textwidth}
		\begin{tabular}{l|c|c|c|c}
			\hline
			\multirow{2}{*}{\textbf{Model}} & \multicolumn{4}{c}{\textbf{DSTC4}}\\ \cline{2-5}
			& \textbf{Slot (F1)}      & \textbf{Intent (F1)}   & \textbf{Intent (Acc)}   & \textbf{Overall (Acc)} \\ \hline
			Attention BiRNN   & 44.0 &  -  & 42.1 & 32.6 \\
			Slot-Gated  & 45.0 &  -  & 42.5 & 32.5 \\
			Slot-gated Intent  & 50.2 &  -  & 40.6 & 31.7 \\
			Bi-Model         & 44.6 &  -  & 41.3 & 30.5 \\
			SF-ID             & 51.4 &  -  & 41.8 & 33.0 \\
			Stack-Propagation (1)  & 52.8 &  -  & 44.9 & 34.6 \\
			Stack-Propagation (2) & 51.9 &  39.2  & 39.2 & 30.5 \\
			\hdashline
			Joint Multiple ID-SF & 48.0 & 37.5 & 39.0 & 29.4 \\  \hline
			AGIF         & \textbf{53.9} & \textbf{40.0} & \textbf{46.1} & \textbf{35.2} \\ \hline
		\end{tabular}
	\end{adjustbox}
	\caption{Slot filling and intent detection results on the DSTC4 dataset. Stack-Propagation (1) denotes the Stack-Propagation (\textit{concatenation}) version and Stack-Propagation (2) denotes the Stack-Propagation (\textit{sigmoid-decoder}) version.}
	\label{tab:multi_intent_results_dstc4}
\end{table}

From the results, we have three observations: 

\begin{figure*}[h]
	\centering
	\includegraphics[width=1.0\textwidth]{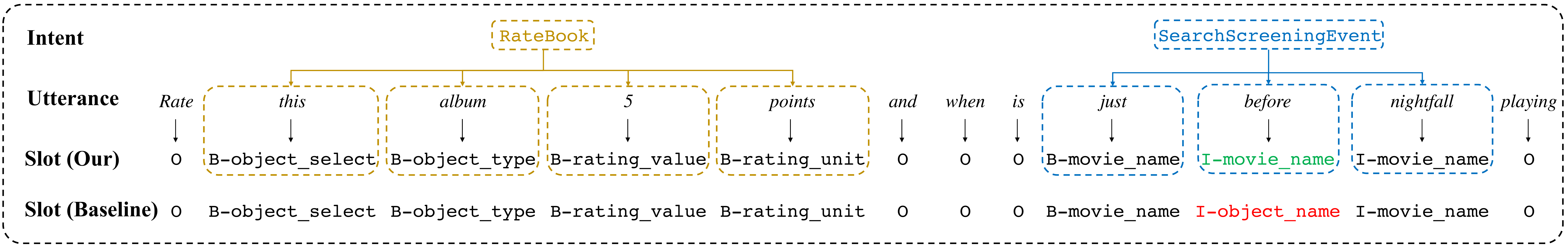}
	\caption{
A case study between our model and \textit{Joint Multiple ID-SF}. The green slot is correct while the red one is wrong. Better viewed in color.		
	}. 
	\label{fig:case-study}
\end{figure*}

1) Our framework outperforms \textit{Joint Multiple ID-SF} baseline by a large margin and achieves state-of-the-art performance.
On the MixATIS dataset, we achieve 0.6\% improvement on Slot (F1) score, 0.6\% improvement on Intent (F1), 2.7\% improvement on Intent (Acc).
On the MixSNIPS dataset, we achieve 3.5\% improvement on Slot (F1) score, 0.4\% improvement on Intent (F1), 0.8\% improvement on Intent (Acc). 
This indicates that our adaptive intent-slot graph interaction successfully incorporates relevant intent information to improve slot prediction.
In addition, we obtain 6.4\% improvement and 9.8\% improvement on Overall (Acc) on MixATIS and MixSNIPS dataset, respectively.
We attribute this to the fact that our adaptive intent-slot graph interaction mechanism can better help grasp the relationship between the intent and slots and improve the whole SLU.

2) 
The \textit{concatenation} outperforms the \textit{sigmoid-decoder} version, this is because \textit{concatenation} can greatly reduce the multi-intent search space, which makes it easier for single intent systems to predict multiple intents.
For example, on the ATIS dataset, there exist 17 single intents and 4 combined multi-intent in the training data. The multi-intent systems make a binary prediction at each intent while the \textit{concatenation} model predicts the limited combined intent search space $(17+4)$.

3) 
Though facing the difficulty of multi-intent prediction, 
our framework outperforms the state-of-the-art single-intent model (\textit{Stack-Propagation (concatenation)}), which further proves the proposed token-level adaptive graph interaction layer can improve the SLU performance. 

\subsection{Analysis}
\subsubsection{Performance on the DSTC4 dataset}
To further analyze the performance of the AGIF model, we conduct experiments on the real-world multi-intent SLU dataset, DSTC4. The results are shown in Table~\ref{tab:multi_intent_results_dstc4}. From the results, we achieve 5.9\% improvement on Slot (F1) score, 2.5\% improvement on Intent (F1), 7.1\% improvement on Intent (Acc) and 5.8\% improvement on Overall (Acc) compared with \textit{Joint Multiple ID-SF}. This further proves that our adaptive intent-slot graph interaction could aggregate the pertinent intent information to enhance the token-level slot prediction.

\begin{table}[tbp]
	\centering
	\begin{adjustbox}{width=0.48\textwidth}
		\begin{tabular}{l|c|c|c|c}
			\hline
			\multirow{2}{*}{\textbf{Model}} & \multicolumn{4}{c}{\textbf{MixSNIPS}} \\ \cline{2-5}
			& \textbf{Slot (F1)}      & \textbf{Intent (F1)}   & \textbf{Intent (Acc)}   & \textbf{Overall (Acc)} \\ \hline
			Vanilla Attention Interaction & 93.8 & 98.0 & 95.2 & 74.0 \\
			GCN-based Interaction & 93.3 & 98.3 & 96.0 & 72.7 \\
			Sentence-Level Augmented & 93.8 & 98.1 & 95.7 & 73.9 \\ 
			\quad+ More Parameters  & 94.1 & 98.6 & 96.6 & 73.6 \\ 
			\hline
			AGIF & \textbf{94.5} & \textbf{98.6} & \textbf{96.5} & \textbf{76.4} \\ \hline
			
		\end{tabular}
	\end{adjustbox}
	\caption{Ablation Study on MixSNIPS Datasets.}
	\label{tab:ablation}
\end{table}

\subsubsection{Effectiveness of Intent-Slot Graph Interaction Mechanism}
\begin{itemize}
	\item {\bf{Graph Attention Mechanism vs. Vanilla Attention Mechanism}} Instead of adopting the GAT to model the interaction between the predicted intents and slot, we utilize the attention mechanism to incorporate the intents information for slot filling at the token-level. We name it as \textit{Vanilla Attention Interaction}.
	We first use the hidden state of slot filling decoder as the query to attend to the intent embedding to obtain the context intent vector, and then we sum the vector and the hidden state of slot filling decoder to get the final slot prediction.
	The results are shown in \textit{Vanilla Attention Interaction} row in Table~\ref{tab:ablation}, we observe the overall 
	performance drops 2.4\% on the MixSNIPS dataset. 
	We attribute it to the fact that the multi-layer graph attention network can automatically capture relevant intents information and better aggregate intents information for each token slot prediction.
	
	\begin{figure}[tbp]
		\centering
		\includegraphics[width=0.48\textwidth]{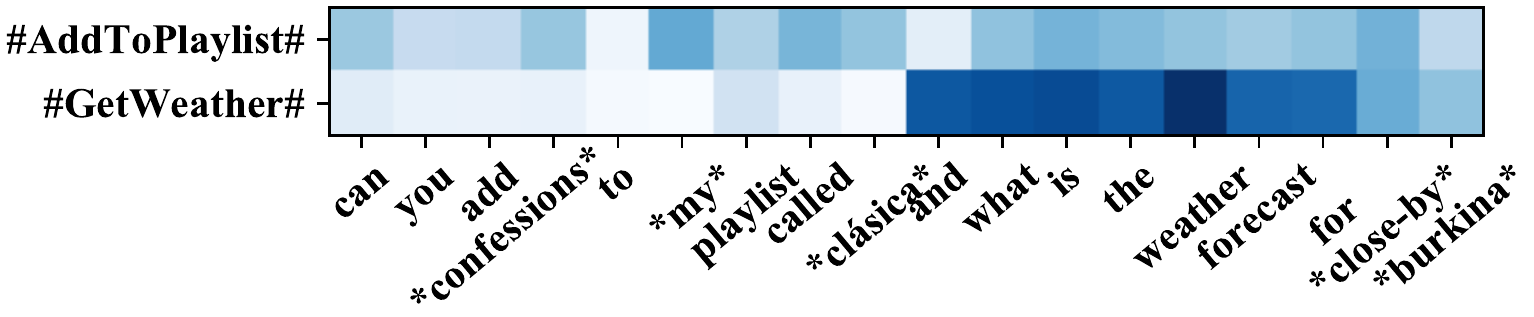}
		\caption{Visualization.
			Y-axis is the predicted intents and X-axis is the input utterance where slot tokens are surrounded by $*$. For each column, the darker the color, the more relevant they are.
		}
		\label{fig:visualization}
	\end{figure}

	\item {\bf{Graph Attention Mechanism vs. Graph Convolution Mechanism}}  We replace the graph attention layer with the graph convolution layer and keep other components unchanged.
	We refer to it as \textit{GCN-based Interaction}.
	The results are shown in \textit{GCN-based Interaction} row in Table~\ref{tab:ablation}, we observe the performance drops in all metrics in the MixSNIPS dataset. 
	We suggested that \textit{GCN-based Interaction} cannot adaptively attribute different weights to each node in the intent-slot graph while our graph attention mechanism can automatically filter irrelevant intent information for each token.
\end{itemize}

\begin{table*}[tbp]
	\centering
	\begin{adjustbox}{width=1.0\textwidth}
		\begin{tabular}{l|c|c|c|c|c|c|c|c}
			\hline
			\multirow{2}{*}{\textbf{Model}} & \multicolumn{4}{c|}{\textbf{ATIS}} & \multicolumn{4}{c}{\textbf{SNIPS}} \\ \cline{2-9}
			& \textbf{Slot (F1)}      & \textbf{Intent (F1)}   & \textbf{Intent (Acc)}   & \textbf{Overall (Acc)}& \textbf{Slot (F1)}      & \textbf{Intent (F1)}   & \textbf{Intent (Acc)} & \textbf{Overall (Acc)}   \\ \hline
			SF-ID               & 95.6  & -  & 96.6 & 86   & 90.5  & -  & 97    & 78.4 \\
			Stack-Propagation      & 95.9  & -  & 96.9 & 86.5 & 94.2  & -  & 98.0  & 86.9 \\
			Joint Multiple ID-SF
			& 94.2  & -  & 95.4 & -    & 88.0  & -  & 97.2  & -    \\ \hline
			AGIF & \textbf{96.0} & \textbf{80.2} & \textbf{97.1} & \textbf{87.2} & \textbf{94.8} & \textbf{98.3} & \textbf{98.1} & \textbf{87.3} \\ \hline
		\end{tabular}
	\end{adjustbox}
	\caption{Slot filling and intent detection results on two single-intent datasets. }
	\label{tab:single_intent_results}
\end{table*}

\subsubsection{Effectiveness of Adaptive Intent-Slot Interaction Mechanism}
\begin{itemize}
	\item {\bf{Adaptive Interaction Mechanism vs. Sentence-Level Augmented Mechanism}} We first conduct experiments by statically providing the same intent information for all tokens slot prediction where we sum the predicted intent embeddings and directly add it to the hidden state of slot filling decoder.
	We refer to it as \textit{sentence-level augmented}.
	The result is shown in Table~\ref{tab:ablation}.
	We can observe that if we only provide overall intent information for slot filling, we obtain the worse results, which demonstrates the effectiveness of adaptively incorporating intent information at the token-level.
	We believe the reason is that 
	providing the same intents for all tokens can cause the ambiguity where each token is hard to extract the relevant intent information while our adaptive intent interaction mechanism can achieve the fine-grained intent interaction and capture the related intent information to guide the slot prediction.

	A natural question that raised is whether the more parameters involved by AGIF contribute to the final performance.
	To verify that the proposed adaptive interaction mechanism rather than the added parameters works, for 
	\textit{sentence-level augmented mechanism} model, we apply multiple LSTM layers (2-layers) to slot filling decoder  
	and we name it as \textit{more parameters}.
	The results in Table~\ref{tab:ablation} show that our framework outperforms the \textit{more parameters} model in overall accuracy, which verifies that the improvements comes from the proposed adaptive intent-slot interaction mechanism rather than the involved parameters. 

	\item {\bf{Qualitative Analysis.}} We provide a case study to intuitively understand the  token-level adaptive intent-slot interaction mechanism. 
	As shown in Figure~\ref{fig:case-study}, AGIF predicts ``\texttt{I-movie\_name}'' correctly  for the slot label of ``\textit{before}'' while \textit{Joint Multiple ID-SF} predicts it as ``\texttt{I-object\_name}'' incorrectly. 
	We observed that ``\texttt{I-object\_name}'' doesn't belong to the intent ``\texttt{SearchScreeningEvent}'' but to the intent ``\texttt{RateBook}''. We attribute it to the reason that each token is guided with the same complex intents information making it incorrectly and confusedly capture the information of the other intent ``\texttt{RateBook}''. 
	In contrast, our adaptive graph interaction mechanism can offer fine-grained intent information integration for token-level slot filling to predict the slot label correctly.
\end{itemize}

\subsubsection{Visualization}
With the attempt to better understand what the adaptive intent-slot graph interaction layer has learned, 
we visualize the intent attention weights of slot filling hidden states node in the output head of the adaptive intent-slot graph interaction layer, which is shown in Figure~\ref{fig:visualization}.
Based on the utterance ``\textit{can you add confessions to my playlist called clásica and what is the weather forecast for close-by burkina}'' and the intents ``\texttt{AddToPlaylist}'' and ``\texttt{GetWeather}'', we can clearly see the attention weights successfully focus on the correct intent, which means our graph interaction layer can learn to incorporate the correlated intent information at each slot. More specifically, our model properly aggregates the corresponding ``\texttt{AddToPlaylist}'' intent information at slots ``\textit{confessions, my, clásica}'' and ``\texttt{GetWeather}'' intent information at slots``\textit{close-by burkina}''. 

\subsubsection{Evaluation on the Single-Intent Datasets}
\label{singleresult}

We conduct experiments on two public single-intent benchmarks to evaluate the generalizability of our framework.
We compare our model with the single-intent state-of-the-art models including \textit{SF-ID}, \textit{Stack-Propagation} and multi-intent model including \textit{Joint Multiple ID-SF}.
 Table~\ref{tab:single_intent_results} shows the experiment results of the proposed models on the ATIS and SNIPS datasets.
From the table, we can see that our model outperforms all the compared baselines and achieves state-of-the-art performance.
This demonstrates the generalizability and effectiveness of our framework whether handling 
multi-intent or single-intent SLU.

\section{Related Work}
\label{RelatedWork}
\paragraph{Intent Detection}
Intent detection is formulated as an utterance classification problem.
Different classification methods, such as support vector machine (SVM) and RNN \citep{haffner2003optimizing,sarikaya2011deep}, have been proposed to solve it. \citet{xia-etal-2018-zero} adopts a capsule-based neural network with self-attention for intent detection. However, the above models mainly focus on the single intent scenario, which can not handle the complex multiple intent scenario.  \citet{Xu2013Exploiting} and \citet{kim2017two} explore the complex scenario, where multiple intents are assigned to a user’s utterance. \citet{Xu2013Exploiting} use log-linear models to achieve this, while we use neural network models. 
Compared with their work, we jointly perform multi-label intent detection and slot prediction,
while they only consider the subtask intent detection.
\paragraph{Slot Filling}
Slot filling can be treated as a sequence labeling task.
The popular approaches are conditional random fields (CRF) \citep{raymond2007generative} and recurrent neural networks (RNN) \citep{xu2013convolutional,yao2014spoken}. 
Recently, \citet{shen2018disan} and \citet{tan2018deep} introduce the self-attention mechanism for CRF-free sequential labeling.
\paragraph{Joint Model}
To consider the high correlation between intent and slots, many joint models \citep{goo2018slot,li2018self,xia-etal-2018-zero,e-etal-2019-novel,liu2019cm,qin-etal-2019-stack} are proposed to solve two tasks.
\citet{goo2018slot, li2018self,zhang-etal-2019-joint} propose to utilize the intent information to guide the slot filling.
\citet{qin-etal-2019-stack} further utilize a stack-propagation framework for better leveraging intent semantic information to guide the slot filling, which achieves the state-of-the-art performance.
\citet{wang2018bi} and \citet{e-etal-2019-novel} consider the cross-impact between the slot and intents.
Our framework follows those state-of-the-art joint model paradigm, and further focus on the multiple intents scenario while the above joint models do not consider. 
Recently,
\citet{gangadharaiah2019joint} propose a joint model to consider the multiple intent detection and slot filling simultaneously where they explicitly leverage overall intent information with the gate mechanism to guide all tokens slot prediction.
Compared with this work, the main differences are as following:
1) Our framework exploits a fine-grained intent information transfer with a unified graph interaction architecture while their work simply incorporates the same intents information for all tokens slot prediction.
2) As far as we know, their corpus and code are not distributed, which makes it hard to follow. In contrast, we empirically construct two large-scale multi-intent SLU datasets where all datasets and code have been released. We hope it would push forward the research of multi-intent SLU.

\section{Conclusion}
\label{Conclusion}
In our paper, we propose a token-level adaptive graph-interactive framework to model the interaction between multiple intents and slot at each token, which can make a fine-grained intent information transfer for slot prediction.
To our best of knowledge, this is the first work to explore fine-grained intents information transfer in multi-intent SLU.
In addition, we release two multi-intent datasets and hope it can push forward the research this area.
Experiments on four datasets show the effectiveness of the proposed models and achieve state-of-the-art performance. 

\section*{Acknowledgements}
We thank the anonymous reviewers for their helpful comments and suggestions.
This work was supported by the National Natural Science Foundation of China (NSFC) via grant 61976072, 61632011 and 61772153. 

\bibliography{emnlp2020}
\bibliographystyle{acl_natbib}

\end{document}